\documentclass{article}
\pdfoutput=1
\usepackage{arxiv}

\usepackage[utf8]{inputenc} 
\usepackage[T1]{fontenc}    
\usepackage{hyperref}       
\usepackage{url}            
\usepackage{booktabs}       
\usepackage{amsfonts}       
\usepackage{nicefrac}       
\usepackage{microtype}      
\usepackage{lipsum}

\usepackage{amsmath}
\usepackage{graphicx,psfrag,epsf}
\usepackage{enumerate}
\usepackage{natbib}

\usepackage{adjustbox}
\usepackage{multirow}
\usepackage{xcolor}
\usepackage{caption}
\usepackage{subcaption}
\usepackage{chngcntr}
\usepackage[flushleft]{threeparttable}

\renewcommand{\theequation}{\arabic{equation}}

\title{\bf DeepCENT: Prediction of Censored Event Time via Deep Learning}

\author{
Yichen Jia\\
Department of Biostatistics\\
Graduate School of Public Health\\
University of Pittsburgh, Pittsburgh, USA\\
\And
Jong-Hyeon Jeong\thanks{Corresponding Author: jjeong@pitt.edu}\\
Department of Biostatistics\\
Graduate School of Public Health\\
University of Pittsburgh, Pittsburgh, USA\\
}

\begin{document}
	\maketitle

\begin{abstract}
With the rapid advances of deep learning, many computational methods have been developed to analyze nonlinear and complex right censored data via deep learning approaches. However, the majority of the methods focus on predicting survival function or hazard function rather than predicting a single valued time to an event. In this paper, we propose a novel method, DeepCENT, to directly predict the individual time to an event. It utilizes the deep learning framework with an innovative loss function that combines the mean square error and the concordance index. Most importantly, DeepCENT can handle competing risks, where one type of event precludes the other types of events from being observed. The validity and advantage of DeepCENT were evaluated using simulation studies and illustrated with three publicly available cancer data sets.
\end{abstract}

\section{Introduction}

Survival analysis, or more generally time-to-event analysis, typically refers to a statistical branch to link a vector of covariates to a summary measure of the time-to-event distribution. Two fundamental functions in survival analysis are the survival function and hazard function. Survival function, denoted as $S(t) = P(T > t)$, is the probability of survival beyond time $t$. On the other hand, the hazard function is a measure of instantaneous failure rate at time $t$, conditional on having survived prior to time $t$,
$$h(t) = \lim\limits_{\triangle t \to 0} \frac{P(t\leq T < t + \triangle t | T \geq t)}{\triangle t}. $$ The key feature of survival analysis is censoring, i.e. the event of interest not occurring during the study period. Therefore, the special methods like the Kaplan-Meier \citep{kaplan1958nonparametric} or Cox proportional hazard methods \citep{cox1975partial} have been developed to accommodate the censoring mechanism into the analysis.

In practice, we often encounter competing risks data, which is a special type of survival data where a competing event preludes the primary event of interest from being observed. For example, in cancer research, if cancer-related mortality is of primary interest, then deaths from other causes can be considered as competing events. Statistical methods for competing risks data have been well established based on cause-specific hazards \citep{kalbfleisch2011statistical} or cumulative incidence function  approaches \citep{pepe1993kaplan}.

With rapid advances in the area of deep learning, there are more flexible alternatives for analyzing nonlinear and complex right censored data. However, rather than focusing on predicting an individual time to an event, the deep learning algorithms for survival analysis focus more on predicting a summary measure of the true failure time distribution, such as survival function, hazard function, or the quantile survival time with or without covariate adjustments. For example, several deep learning frameworks in survival analysis utilized the Cox model-based loss function, which predicts the patient's risk \citep{faraggi1995neural,katzman2018deepsurv,ching2018cox}. Alternatively, discrete-time survival model with deep learning framework has been also developed to predict survival probability \citep{fotso2018deep, giunchiglia2018rnn}, which has been extended to a competing risks setting \citep{lee2018deephit}. A deep learning method for quantile regression on right censored time has also been proposed so that the quantile survival time like the median survival time can be predicted \citep{jia2022deep}.

An individual patient might be interested in knowing how long (s)he would live or prolong an onset of a disease if (s)he receives a existing or new therapy, i.e. estimated individual survival time given their genetic or/and environmental backgrounds (Henderson and Keiding, 2005). To directly predict individual's survival time, one can use a parametric approach where the underlying distribution is pre-specified, but it would not be robust to a model misspecification. In the deep learning framework, \citet{baek2021survival} integrated a pre-trained hazard ratio network (DeepSurv) and a newly proposed distribution function network, and \citet{jing2019deep} utilized a loss function that combined an extended mean squared error (MSE) loss and a pairwise MSE ranking loss, to directly predict individual survival time. To the best of our knowledge, there is no literature on predicting individual survival time under competing risks setting. In this paper, therefore, we developed \textbf{DeepCENT}, a \textbf{Deep} learning algorithm to directly predict an individual's \textbf{C}ensored \textbf{E}ve\textbf{N}t \textbf{T}ime that utilizes an innovative loss function under both noncompeting and competing risks settings. The proposed method has been implemented using \verb!PyTorch! and can utilize graphics processing units (GPU) for acceleration. A detailed tutorial can be found at \url{https://github.com/yicjia/DeepCENT}.

In Section 2, we present the description and implementation of our deep learning method. The proposed algorithm is assessed via simulation studies in Section 3 and illustrated with publicly available data sets in Section 4. Finally, we conclude our paper with a brief discussion in Section 5.

\section{Methods}

\subsection{Architecture}
\begin{figure}
\begin{subfigure}{0.45\textwidth}
  \includegraphics[width=.85\linewidth]{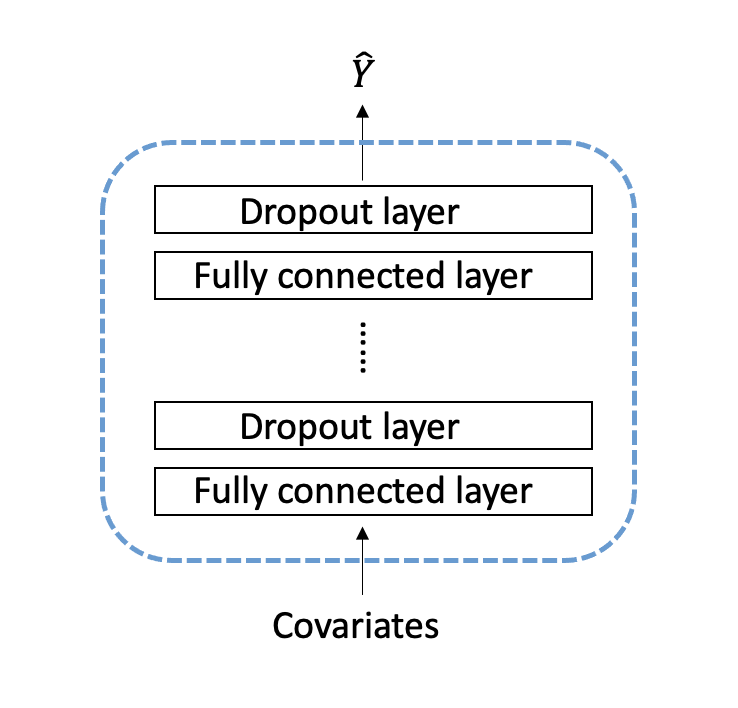}
  \caption{Architecture for survival data without competing risks}
  \label{fig:architecture_regular}
\end{subfigure}
\begin{subfigure}{0.5\textwidth}
  \includegraphics[width=.95\linewidth]{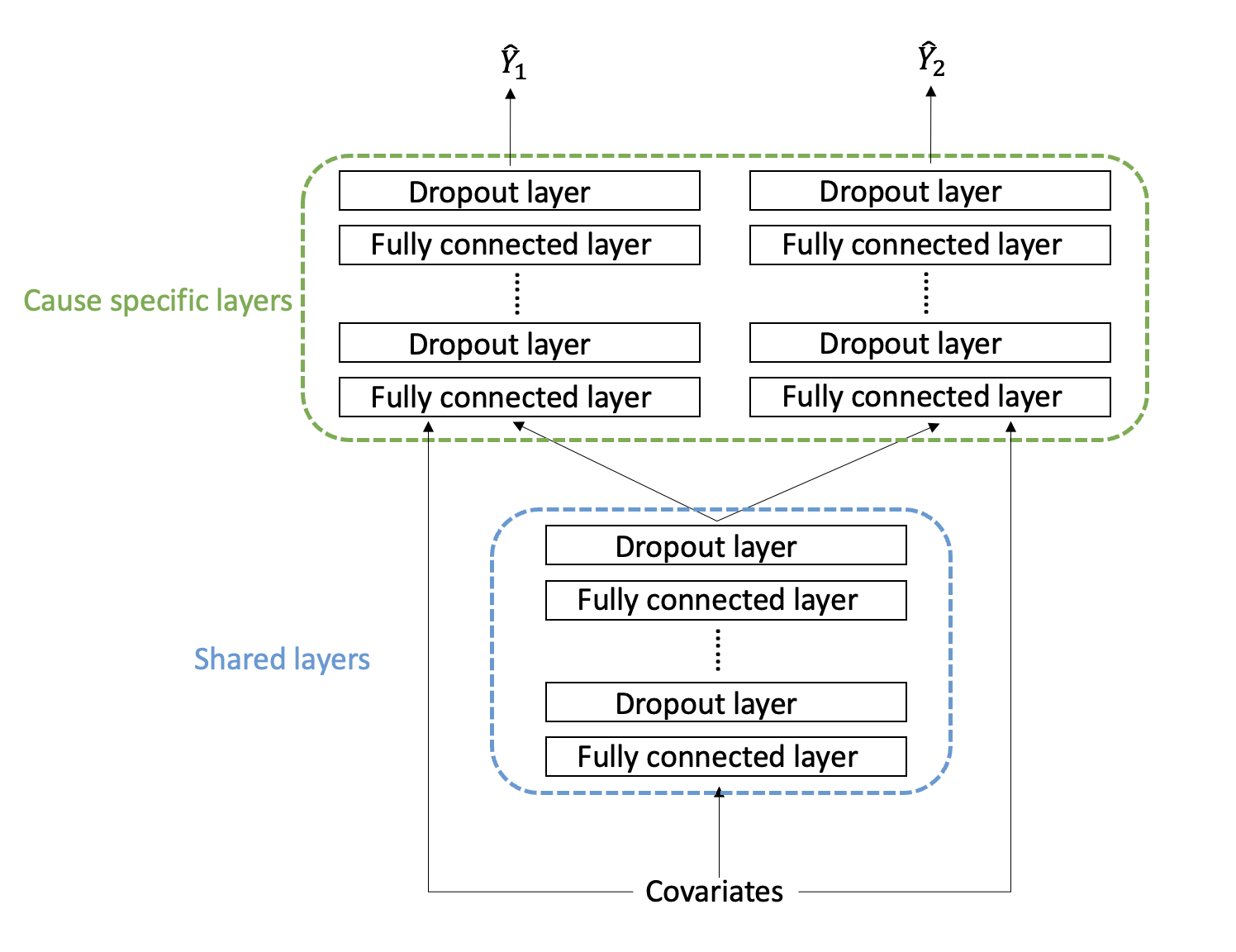}
  \caption{Architecture for competing risks data}
  \label{fig:architecture_cmprsk}
\end{subfigure}
\caption{Architecture of DeepCENT}
\end{figure}

Figure \ref{fig:architecture_regular} shows the architecture of DeepCENT with right censored data without competing risks. We adopt the deep feed-forward neural network to predict individual survival times. The input to the network is a covariate vector. It then consists of several fully connected layers with RELU activation function. Between hidden layers,  dropout layers are also employed to avoid overfitting. The output layer contains a single node for the prediction of an individual survival time.

Under competing risks setting, since we need to predict survival times for multiple events, we adopt the hard parameter sharing structure for DeepCENT. Figure \ref{fig:architecture_cmprsk} shows the architecture of our multi-task neural network \citep{caruana1997multitask} for two competing risks that consists of shared layers where we learn a common representation for all events and additional cause-specific layers that are learned independently for each competing event. Specifically, the output of the shared network, the common representation for all events,
as well as a residual connection \citep{he2016deep} from the original input covariates are the input to the cause-specific sub-networks. Lastly, the output of each cause-specific sub-network is the predicted time to competing events of each type.

\subsection{Loss function}
To train our model, we minimize a total loss function $L_{total}$ that is specifically designed to handle right censored data with or without competing risks. We consider the prediction model as a combination of a regression problem and a ranking problem. Thus, we propose to use a combined loss function that optimizes both regression-based and rank-based objectives simultaneously, which guards against learning models that perform well on one set of metrics but poorly on the other. Specifically, the loss function is the sum of two terms $L_{total} = L_1 + L_2$, where $L_1$ is the regression loss that uses a modified mean square error loss between the predicted and observed survival time that takes censoring into account and $L_2$ is the pairwise ranking loss based on the negative concordance index ($C$-index).

Specifically, under the right censored data without competing risks, let $T_i$ and $C_i$ denote potential failure time and potential censoring time, respectively, and they are independent conditional on the covariate vector $\boldsymbol{x_i}$. In practice, we only observe ($Y_i$, $\delta_i$, $\boldsymbol{x_i}$), where $Y_i = \text{min}(T_i,C_i)$ is the observed survival time, and $\delta_i = I(T_i \leq C_i)$ is the event indicator. Then,

\begin{equation}
\label{mse_loss}
    L_1 = \sum_{i=1}^{n}\{ I(\delta_i = 1)*(\hat{Y}_i - Y_i)^2 + \lambda_1*I(\delta_i = 0)*I(\hat{Y}_i < Y_i)*(\hat{Y}_i - Y_i)^2 \}.
\end{equation}
Note that for observed survival times, i.e. when $I(\delta_i = 1)$, we calculate the regular MSEs. For censored observations, i.e. when $I(\delta_i = 0)$, we include a penalty term when the predicted event time is smaller than the observed censoring time, and the penalty is proportional to the amount of difference. There is no penalty being added if the predicted survival time is greater than the observed censoring time. The tuning parameter, $\lambda_1$, controls the strength of the penalty term, and it is tuned in the hyperparameter tuning step. 

$L_2$ is a ranking loss that utilizes the negative $C$-index, where the $C$-index is a generalization of the area under the ROC curve (AUC) that takes the censoring into account \citep{heagerty2005survival}. It is also related to the rank correlation between the observed and predicted outcomes. Specifically, the $C$-index is the proportion of all comparable pairs by which the predictions are concordant. For example, two samples $i$ and $j$ are comparable if there is an ordering between the two possibly censored outcomes as $Y_i<Y_j \text{ and } \delta_i=1$. Then a comparable pair is concordant if a subject who fails at an earlier time point is predicted with a worse outcome, $\hat{Y}_i < \hat{Y}_j$.
\begin{equation*}
    c = \frac{1}{|\varepsilon|}\sum_{\varepsilon_{ij}}I(\hat{Y}_i < \hat{Y}_j),
\end{equation*}
where $\varepsilon_{ij}$ denotes a comparable pair and $|\varepsilon|$ denotes the total number of comparable pairs.
Since the $C$-index is one of the most popular metrics for evaluating the prediction accuracy for survival data, it is natural to formulate the learning problem to maximize the $C$-index. However, maximizing the $C$-index is a discrete optimization problem, thus we use a modified $C$-index proposed by \citet{steck2008ranking}, where we maximize a differentiable and log-sigmoid lower bound on the indicator function in the $C$-index,
\begin{equation}
\label{rank_loss}
    c^* = \frac{1}{|\varepsilon|}\sum_{\varepsilon_{ij}} 1+(\log\sigma(\hat{Y}_j - \hat{Y}_i)/\log2),
\end{equation}
where $\sigma(z) = 1/(1+e^{-z})$, and hence $\log\sigma$ is a concave log-sigmoid function. It follows easily that $c^*$ is the lower bound of $c$,
\begin{equation*}
    c = \frac{1}{|\varepsilon|}\sum_{\varepsilon_{ij}}I(\hat{Y}_j - \hat{Y}_i > 0) \geq \frac{1}{|\varepsilon|}\sum_{\varepsilon_{ij}} 1+(\log\sigma[\hat{Y}_j - \hat{Y}_i]/\log2).
\end{equation*}

Therefore, $L_2 = -\lambda_2* c^*$ is the negative $C$-index loss and $\lambda_2$ is another tuning parameter. Then, the total loss $L_{total} = L_1 + L_2$, and the optimization of the neural network is achieved by the gradient descent process in the backpropagation phase \citep{rumelhart1986learning} to minimize this total loss function.

Under the competing risks setting, without loss of generality, we assume there are two types of competing events. Let $\epsilon_i = 1$ or 2 indicate the cause of failure, with type 1 events being of primary interest. Then, the event indicator can be defined as $\delta_i = I(T_i \leq C_i)\epsilon_i = 0$, 1 or 2, where $\delta_i = 0$ indicates an independently censored observation. The loss function can be easily extended to
\begin{equation}
\label{mse_loss_cmprsk}
\begin{aligned}
    L_1 = \sum_{i=1}^{n}\{ &I(\delta_i = 1)*(\hat{Y}_{i1} - Y_i)^2 + I(\delta_i = 2)*(\hat{Y}_{i2} - Y_i)^2 + \\
    &\lambda_0* [I(\delta_i = 0)*I(\hat{Y}_{i1} < Y_i)*(\hat{Y}_{i1} - Y_i)^2 + I(\delta_i = 0)*I(\hat{Y}_{i2} < Y_i)*(\hat{Y}_{i2} - Y_i)^2] +\\
    &\lambda_1*I(\delta_i = 1)*I(\hat{Y}_{i2} < \hat{Y}_{i1})*(\hat{Y}_{i2} - Y_{i1})^2 + \lambda_2 *I(\delta_i = 2)*I(\hat{Y}_{i1} < \hat{Y}_{i2})*(\hat{Y}_{i1} - Y_{i2})^2  \},
\end{aligned}
\end{equation}
and
\begin{equation}
\label{rank_loss_cmprsk}
L_2 =-\{ \lambda_3 * \frac{1}{|\varepsilon|}\sum_{\varepsilon_{ij}} 1+(\log\sigma[\hat{Y}_{j1} - \hat{Y}_{i1}]/\log2) + \lambda_4 * \frac{1}{|\varepsilon|}\sum_{\varepsilon_{ij}} 1+(\log\sigma[\hat{Y}_{j2} - \hat{Y}_{i2}]/\log2) \},
\end{equation}
where $\hat{Y}_{i1}$ and $\hat{Y}_{i2}$ denotes the predicted survival time for event types 1 and 2, respectively. Similar to the noncompeting risks setting, the penalty terms contribute only when the predicted survival times for event 1 and 2 are less than the censored observations, and the predicted survival time for type 2 event is less than the predicted survival time for type 1 event when we observe type 1 event, and vice versa.

\subsection{Hyperparameter tuning}
Hyperparameter tuning is crucial in deep learning models to prevent overfitting. In this paper, hyperparameters involved in neural network, including the number of hidden layers (shared layers and cause-specific layers), the number of nodes in each layer, dropout rates, the number of epochs and the tuning parameter $\lambda's$ involved in the loss function, are tuned using the 5-fold cross validation in the training data sets.

\subsection{Prediction interval}
Prediction intervals are also provided using the methods that utilize dropout as Bayesian approximation of the Gaussian process \citep{gal2016dropout}. Originally, dropout was used as a regularization method to avoid over-fitting in neural network by randomly ``dropping out" a portion of nodes output in a given layer only during the training stage \citep{hinton2012improving, srivastava2014dropout}. However, enabling dropout at the test stage and repeating the prediction several times, provides the model uncertainty. One thing to note is that prediction interval measures the model uncertainty in a single predicted outcome, which is different from confidence interval which quantifies the uncertainty in an estimated population parameter.

\section{Simulation Studies}
\label{s:sim}
In this section, we compare our proposed method with the Weibull regression and RankDeepSurv \citep{jing2019deep} under the noncompeting risks setting, and with cause-specific Weibull regression and DeepHit \citep{lee2018deephit} under the competing risks setting.

\subsection{Compared Methods}
\subsubsection{Parametric Weibull Regression}
The Weibull regression,
$$\log(T) = \beta'\mathbf{X} + \sigma \epsilon, $$
where $\epsilon$ follows an extreme value distribution, is the most popular parametric regression model in which the survival time can be estimated as $\exp(\hat{\beta'}\mathbf{X})$.
However, it assumes a log-linear relationship between the covariates and the outcome. Moreover, if the underlying distribution is not a Weibull, then the prediction results can be biased.

\subsubsection{RankDeepSurv}
RankDeepSurv proposed a similar loss function as ours, which also consists of two parts: extended MSE loss and ranking loss. The total loss is $L_{total} = \alpha*L_1 + \beta*L_2$, where $L_1$ is the same as Equation (\ref{mse_loss}) although not including the tuning parameter $\lambda_1$, assuming that the two components contribute equally. The ranking loss, $L_2$, is different from ours in that pairwise MSE loss is calculated between the observed and predicted values of the difference in survival time among all comparable pairs, i.e., $L_2 = \frac{1}{|\varepsilon|}\sum_{\varepsilon_{ij}} [(y_j - y_i) - (\hat{y}_j - \hat{y}_i)]^2$. Even though the pairwise MSE loss is often used in the learning algorithms without censoring, it does include extra information that could result in bias when comparing a pair observations ($y_i,y_j$) with different event indicators in the censored data setting. Specifically, suppose the observation $y_j$ is censored and the observation $y_i$ is an event, then the observed censored time $y_j$ would be a biased measurement for the true event time of subject $j$. On the other hand, the C-index loss implemented in DeepCENT only extracts the ranking information between a pair of predicted outcomes, measuring the concordance among the comparable pairs and hence eliminating the potential bias from using the censored observations in the metric. In addition, the values of $\alpha$ and $\beta$ are pre-fixed in RankDeepSurv and are not tuned in the hyperparameter tuning steps, making DeepCENT more flexible than RankDeepSurv by including the extra tuning parameters. For our comparisons in the simulation study, the code implemented in \citet{jing2019deep} has been rewritten using \verb!PyTorch!.

\subsubsection{DeepHit}
DeepHit is the most popular deep learning algorithm that can handle competing risks. It learns the joint distribution of survival times and competing events directly without making any assumption on the underlying stochastic process. However, DeepHit predicts only the survival probability in pre-specified time intervals. To the best of our knowledge, there is no existing deep learning method that can directly predict individual survival time under the competing risks setting. In this paper, we used the Python package \verb!pycox! to implement DeepHit.

\subsection{Model Evaluation}
The $C$-index and MSE are used for evaluating the prediction performance. Since we know the true survival time in the simulation setting, we compute the MSE between the predicted and true survival time for all observations. However, for competing risks setting, since DeepHit and DeepCENT are predicting different quantities, only the $C$-index is used to assess prediction performance.

It is worth noting that the $C$-index focuses on the order of the predictions while the MSE focuses on the actual deviations of the prediction from the observed outcome. Thus, the $C$-index is not sensitive to small differences in discriminating between two models \citep{harrell1996multivariable}. For example, the $C$-index considers the (prediction, outcome) pairs (0.01, 0), (0.9, 1) as no more concordant than the pairs (0.05, 0), (0.8, 1).

\subsection{Data Generation Mechanism}
Failure time $T$ was generated from an exponential distribution with a scale parameter $\lambda(x)$. To generate nonlinear data, we used the square log-risk function,
$$\lambda(x) = \lambda * \exp(\beta_1*x_1^2 + \beta_2*x_2^2 + \beta_3*x_3^2+ \beta_4*x_4^2),$$
where $\lambda = 2$, $\beta^{\prime}=(\beta_1, \beta_2,\beta_3,\beta_4)= (5, -5, 2, -2)$ and $x^{\prime}=(x_1, x_2,x_3,x_4)$ were generated from $Uniform(-1, 1)$.  A binary predictor $x_5$ following a $Binomial(0.5)$ was included to introduce a multiplicative group effect, with the true failure times for the treatment group ($x_5 = 1$) being a multiple of the true failure times for the control group ($x_5 = 0$). Then, the potential censoring times $C$ were generated from $Uniform(0, \theta)$, where $\theta$ controls the desired censoring proportions as 10\%, 40\%, or 70\%. Finally, the observed survival times were determined as $\min(T,C)$, together with associated event indicator functions, $I(T<C)$.

For competing risks data, both failure time $T_1$ and $T_2$ were generated from exponential distribution with the scale parameter $\lambda_1(x) = \lambda_1*\exp(\beta_1*x_1^2 + \beta_2*x_2^2 + \beta_3*x_3^2)$ and $\lambda_2(x) = \lambda_2*\exp(\beta_1*x_1^2 + \beta_2*x_4^2)$, where $\lambda_1 = 2$, $\lambda_2 = 4$, $\beta = (\beta_1, \beta_2,\beta_3) = (5, -5, 2)$ and $x^{\prime}=(x_1, x_2, x_3, x_4)$ were generated from $Uniform(-1, 1)$. The binary treatment group variable $x_5$ generated from $Binomial(0.5)$ was only associated with type 1 event. The independent censoring times $C$ were generated from $Uniform(0, \theta)$, where $\theta$ controls the overall desired censoring proportions as 10\%, 40\%, or 70\%. The observed survival times were determined by $Y = \min(T_{1},T_{2}, C)$.

For both noncompeting  and competing risks data, we generated training data sets with sample size of $n$ and  test data sets with sample size of $n/2$ where $n = 100, 500 \text{ and } 1000$, and performed 1000 simulations.

\subsection{Results}
\begin{figure}
\begin{subfigure}{0.49\textwidth}
  \centering
  \includegraphics[width=.95\linewidth]{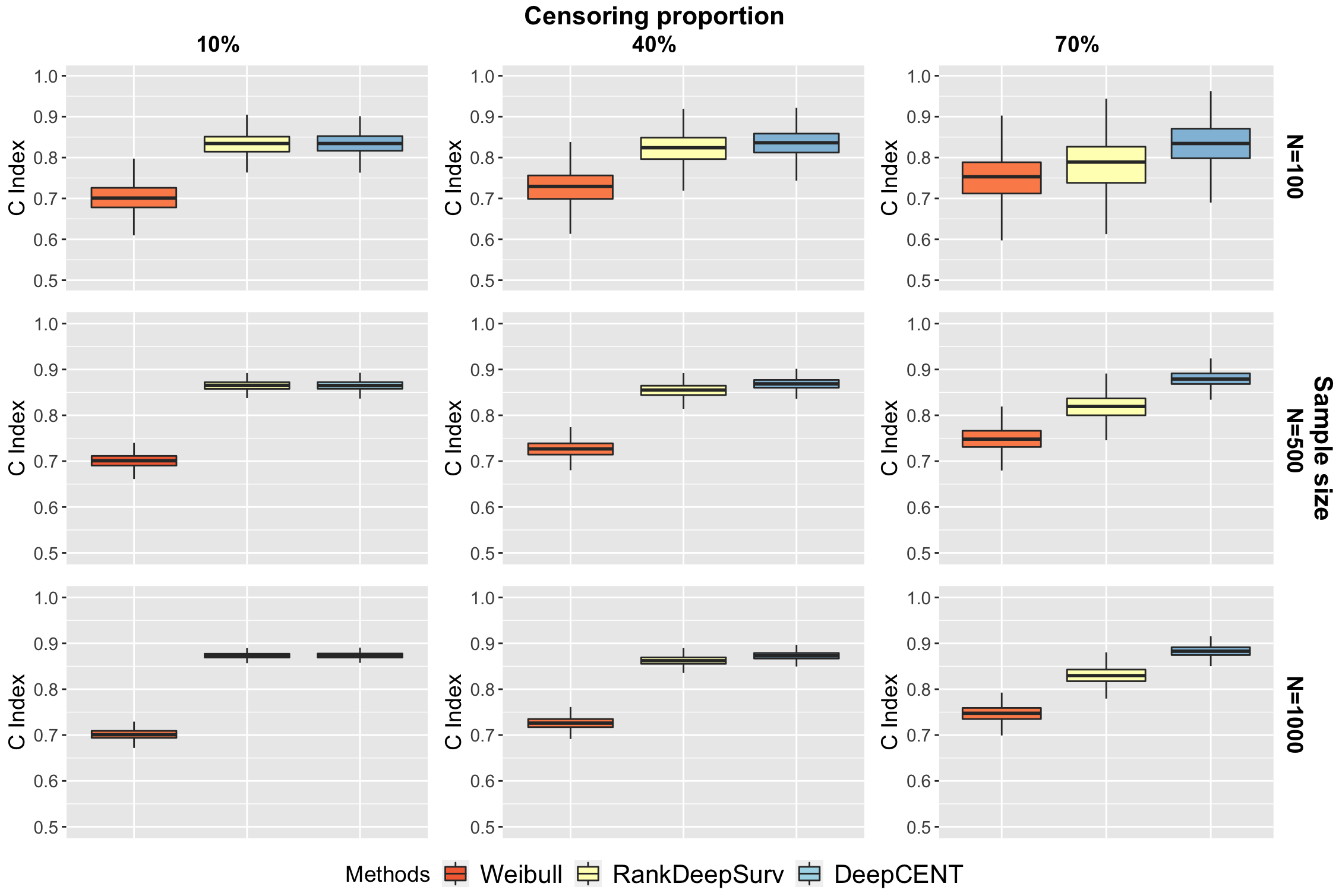}
  \caption{$C$-index}
\end{subfigure}
\begin{subfigure}{0.49\textwidth}
  \centering
  \includegraphics[width=.95\linewidth]{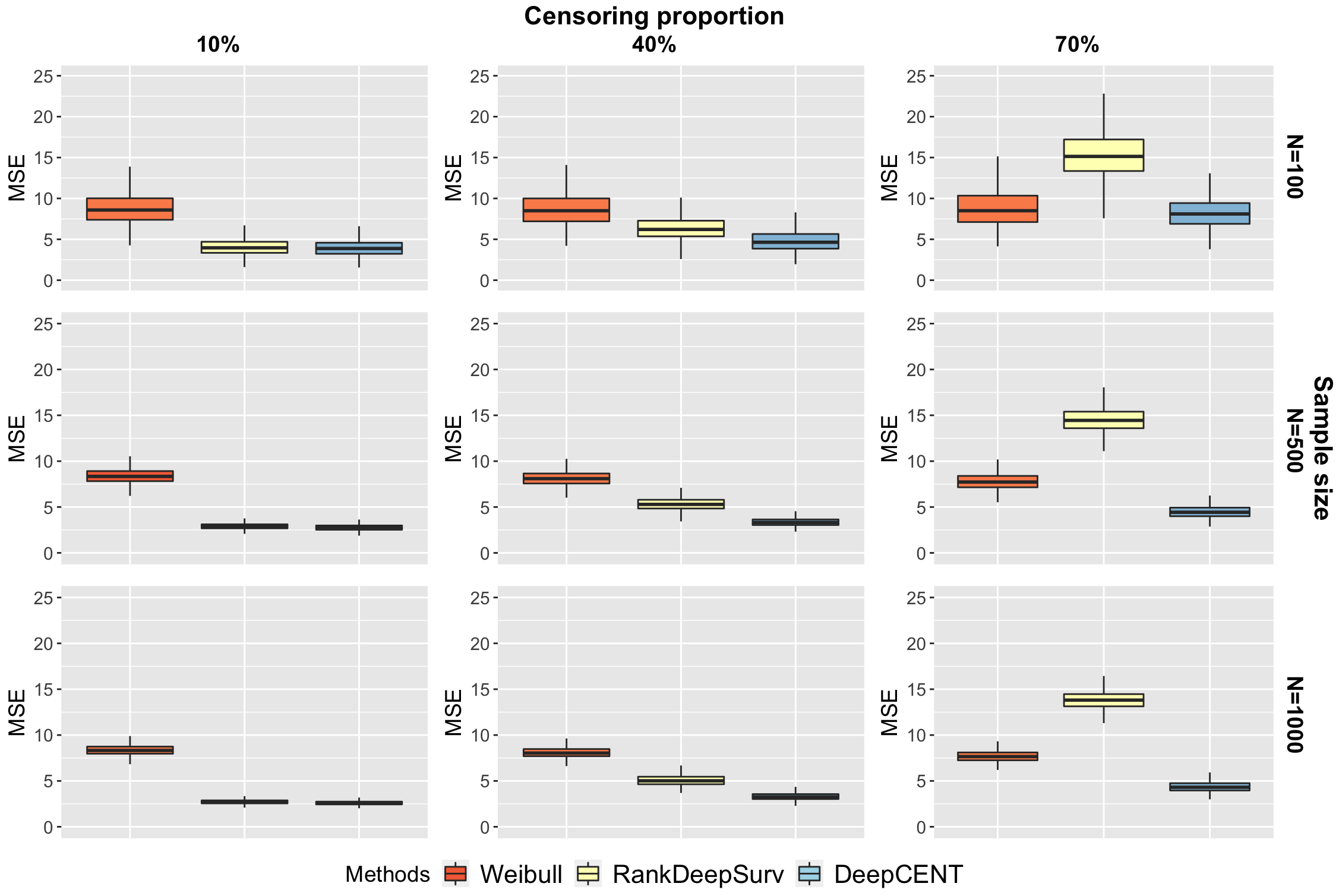}
  \caption{MSE}
\end{subfigure}

\caption{Boxplots of $C$-index and MSE without competing risks}
\label{fig:simulation}
\end{figure}

Figure \ref{fig:simulation} shows the boxplots of the $C$-index and the MSE under different censoring proportions and sample sizes without competing risks. Both DeepCENT and RankDeepSurv outperform the Weibull regression due to the non-linear nature of the simulated data, as expected. DeepCENT performs similarly as RankDeepSurv when the censoring rate is low (e.g., 10\%) but it outperforms RankDeepSurv as the censoring rate increases. As mentioned previously, it is due to the biased information used in the $L_2$ of RankDeepSurv.

\begin{figure}
\begin{subfigure}{0.49\textwidth}
  \centering
  \includegraphics[width=.95\linewidth]{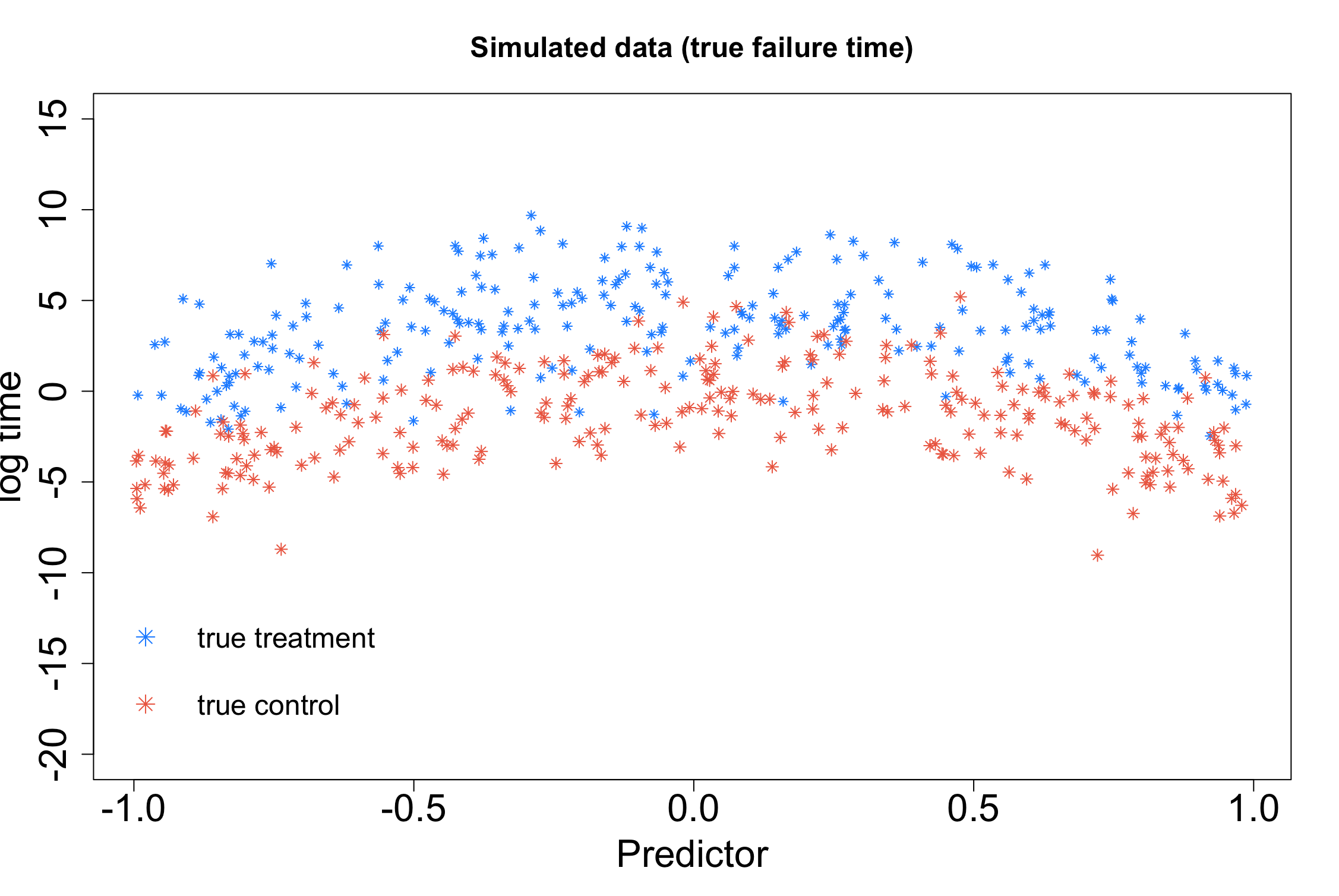}
  \caption{Simulated data (true failure time)}
\end{subfigure}
\begin{subfigure}{0.49\textwidth}
  \centering
  \includegraphics[width=.95\linewidth]{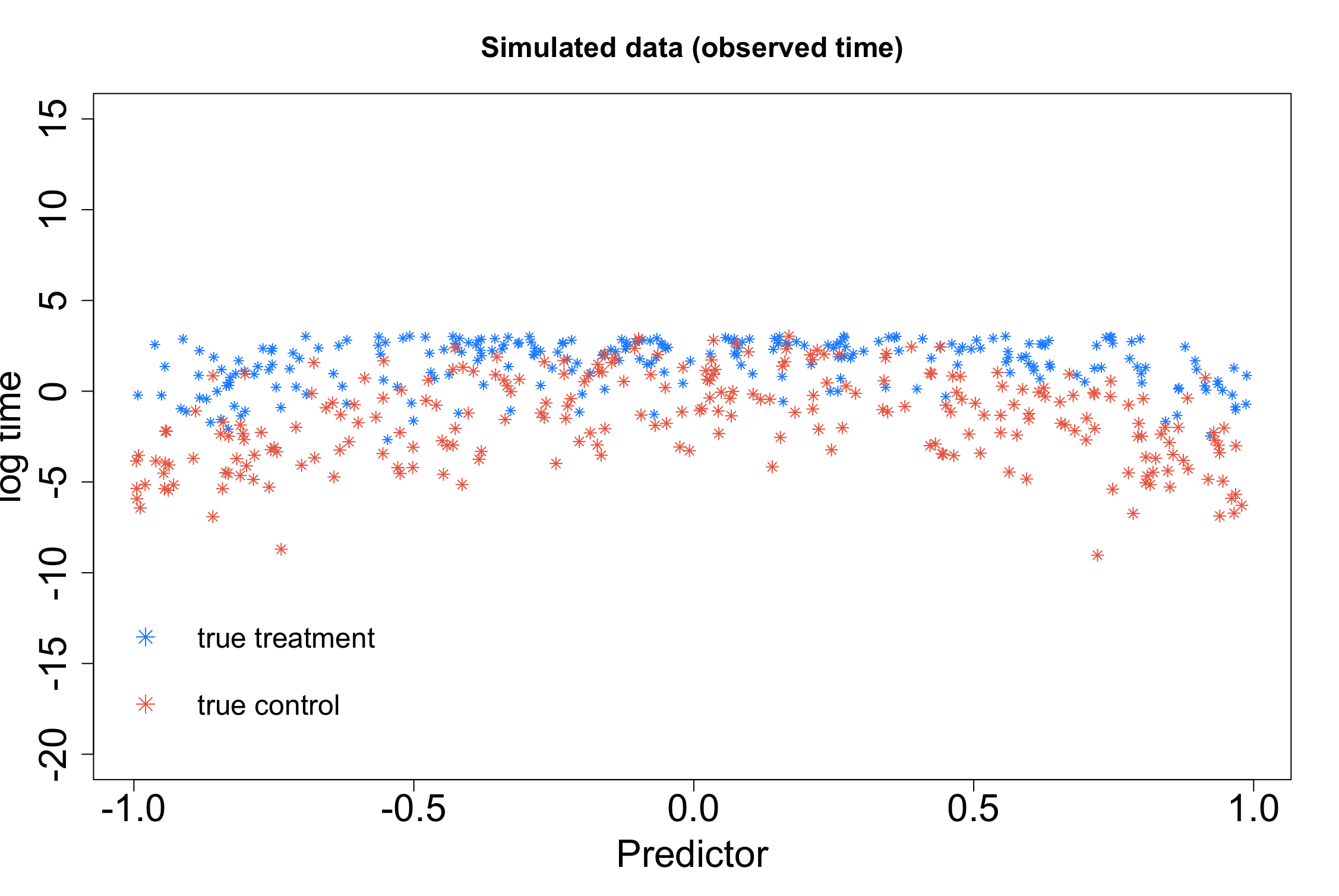}
  \caption{Simulated data (observed time)}
\end{subfigure}

\begin{subfigure}{0.49\textwidth}
  \centering
  \includegraphics[width=.95\linewidth]{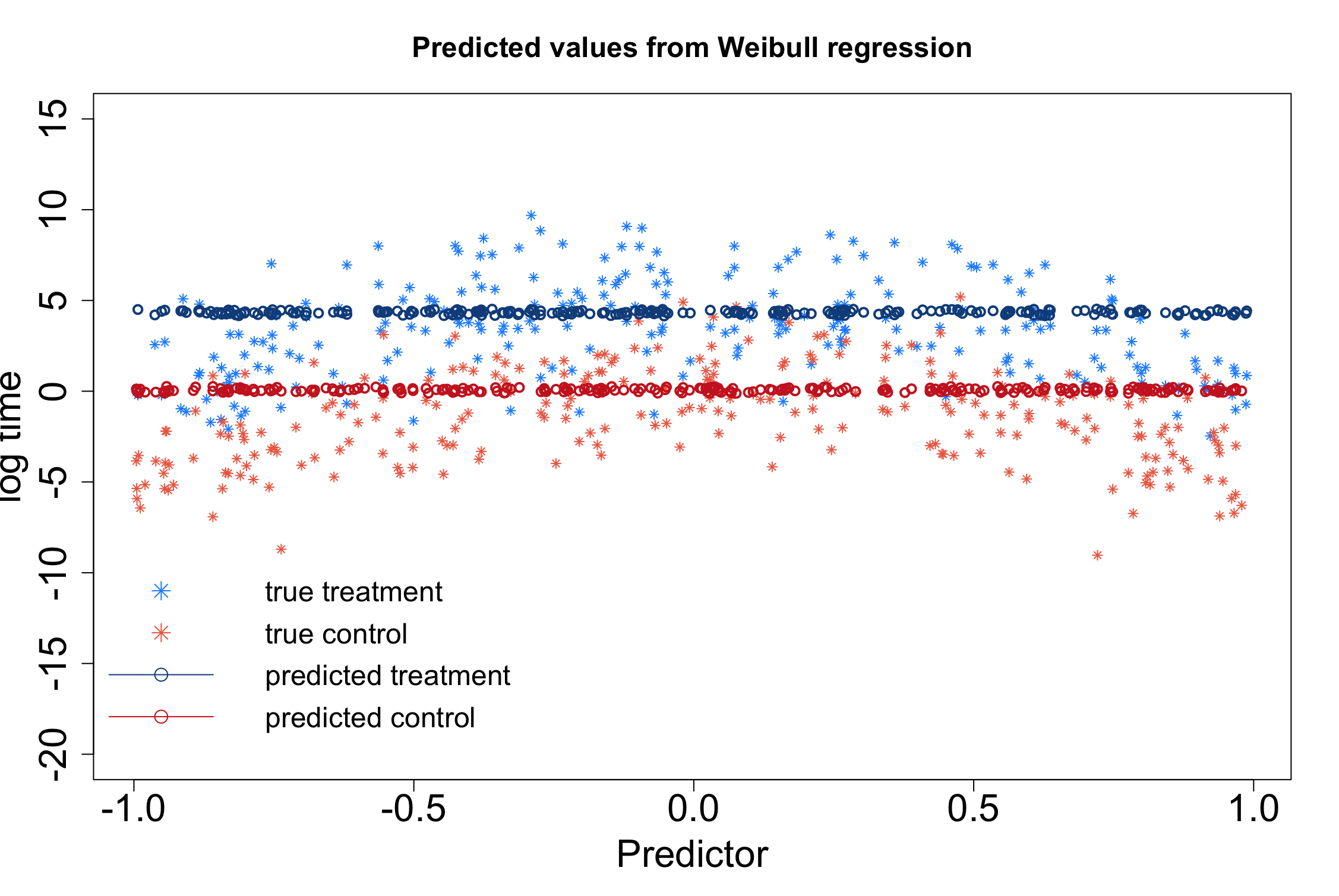}
  \caption{Prediction from Weibull regression}
\end{subfigure}
\begin{subfigure}{0.49\textwidth}
  \centering
  \includegraphics[width=.95\linewidth]{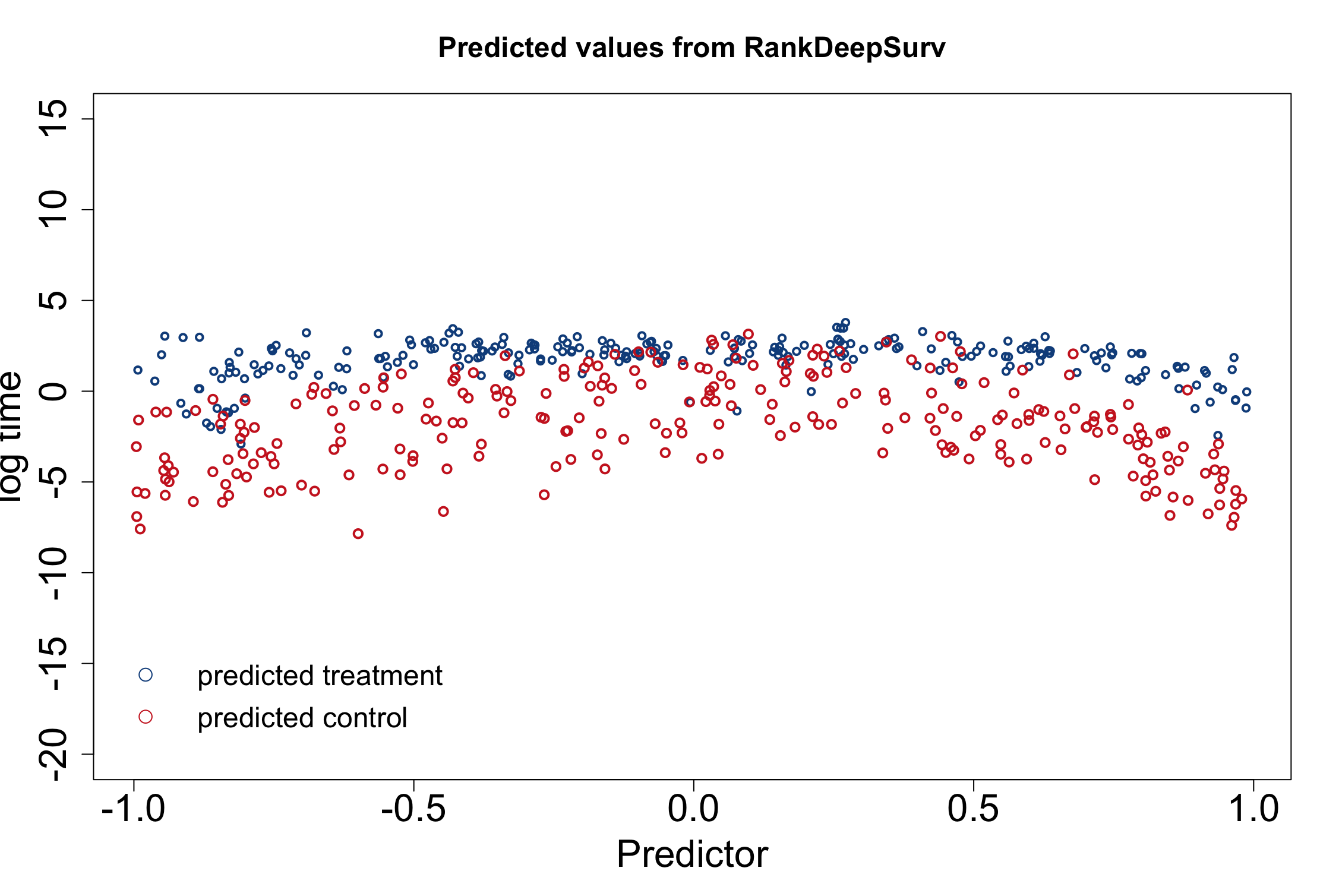}
  \caption{Prediction from RankDeepSurv}
\end{subfigure}

\begin{subfigure}{0.49\textwidth}
  \centering
  \includegraphics[width=.95\linewidth]{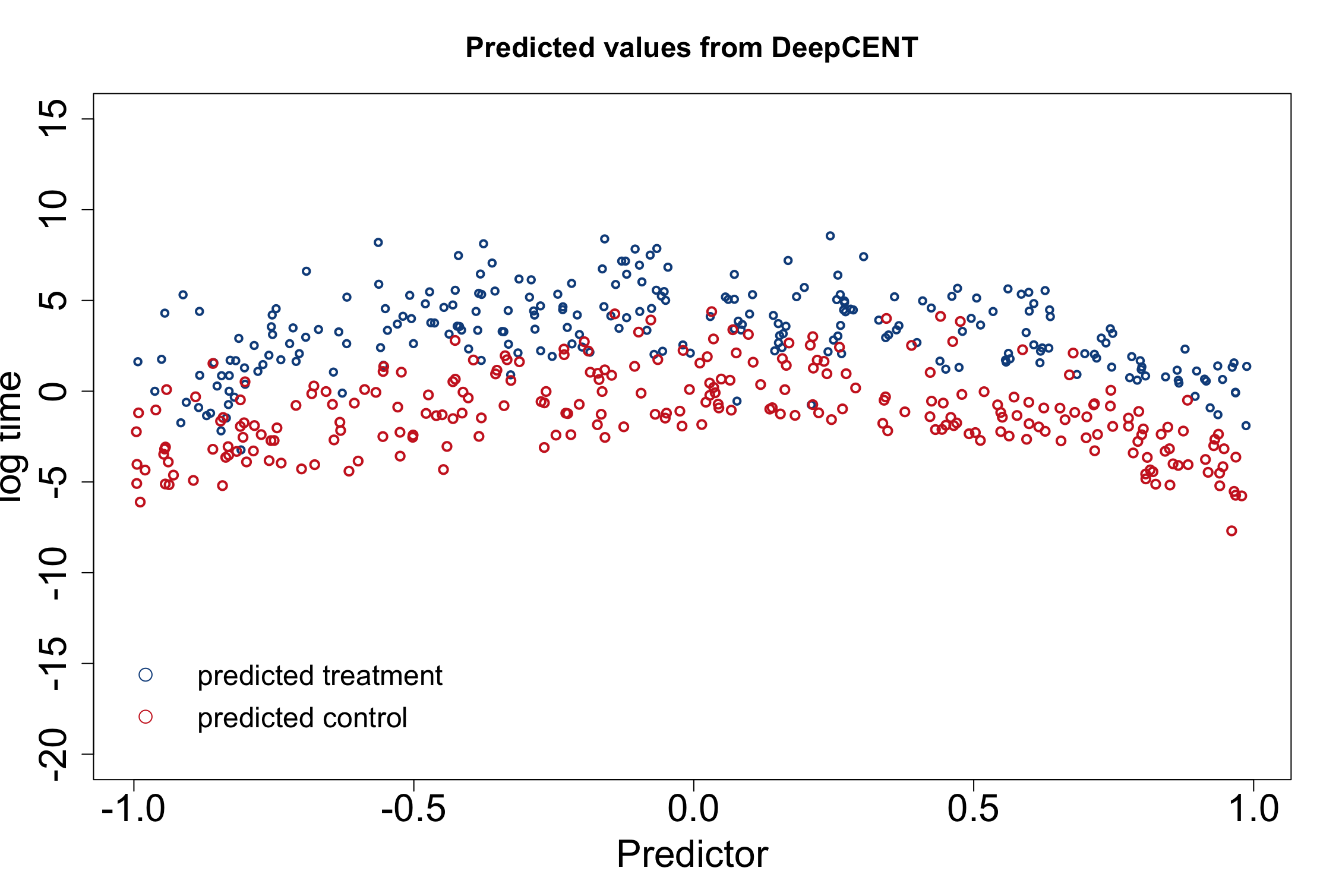}
  \caption{Prediction from DeepCENT}
\end{subfigure}
\caption{Prediction results on a data set with a sample size of $n=500$ and a censoring rate of $= 40\%$}
\label{fig:prediction}
\end{figure}

We also present the prediction results on one of the test data set with a sample size of 500 and a 40\% censoring rate (Figure \ref{fig:prediction}). Both DeepCENT and RankDeepSurv could capture the nonlinear patterns in the data while the Weibull regression model fails to do so due to the linear constraints. The predictions from RankDeepSurv could only recover the observed times since it treats the censoring times as the true failure times in their ranking loss part, $L_2$. On the other hand, the prediction of DeepCENT recovers the true failure times quite well even for those censored observations.

\begin{figure}
\begin{subfigure}{0.49\textwidth}
  \centering
  \includegraphics[width=.95\linewidth]{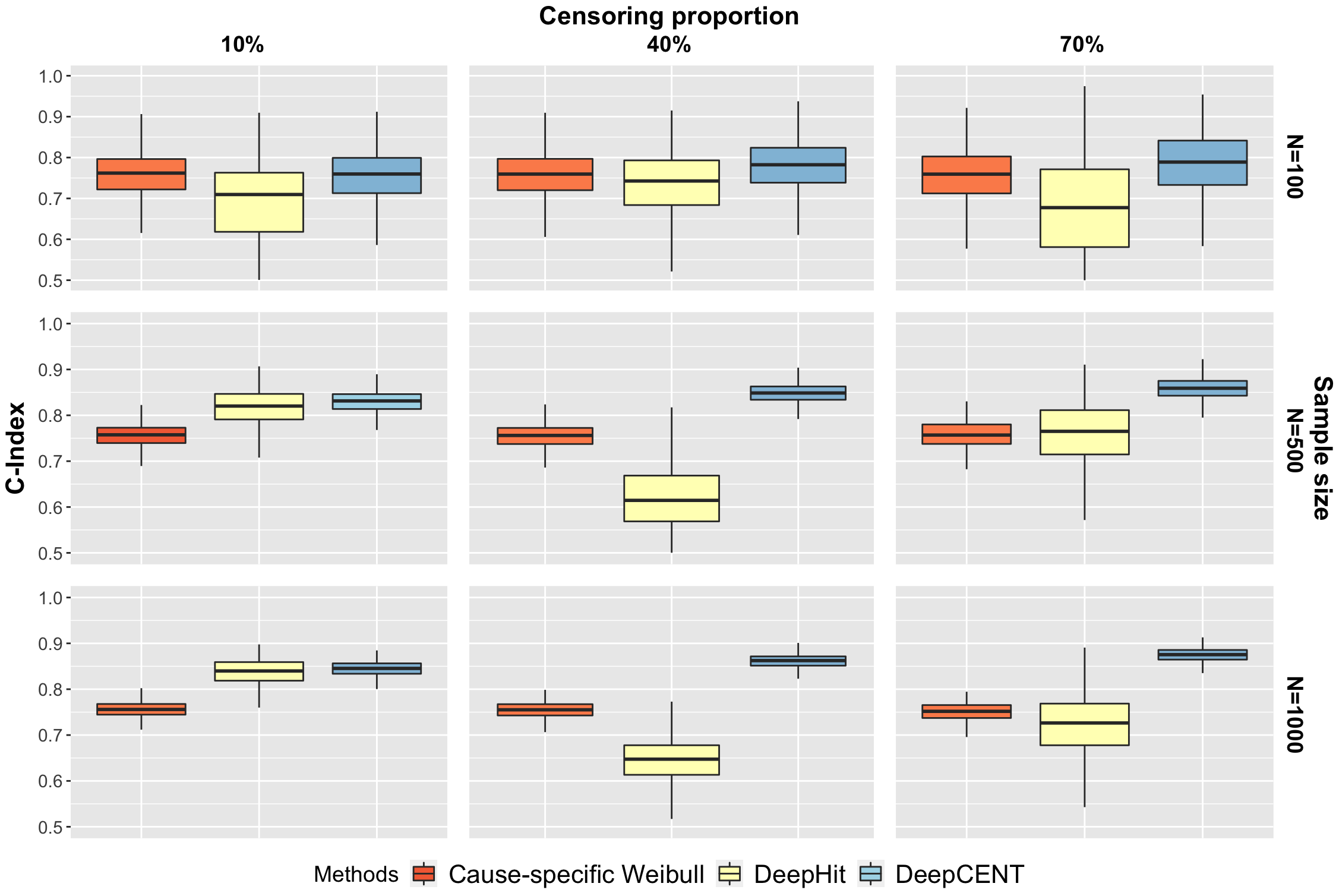}
  \caption{$C$-index for event type 1}
\end{subfigure}
\begin{subfigure}{0.49\textwidth}
  \centering
  \includegraphics[width=.95\linewidth]{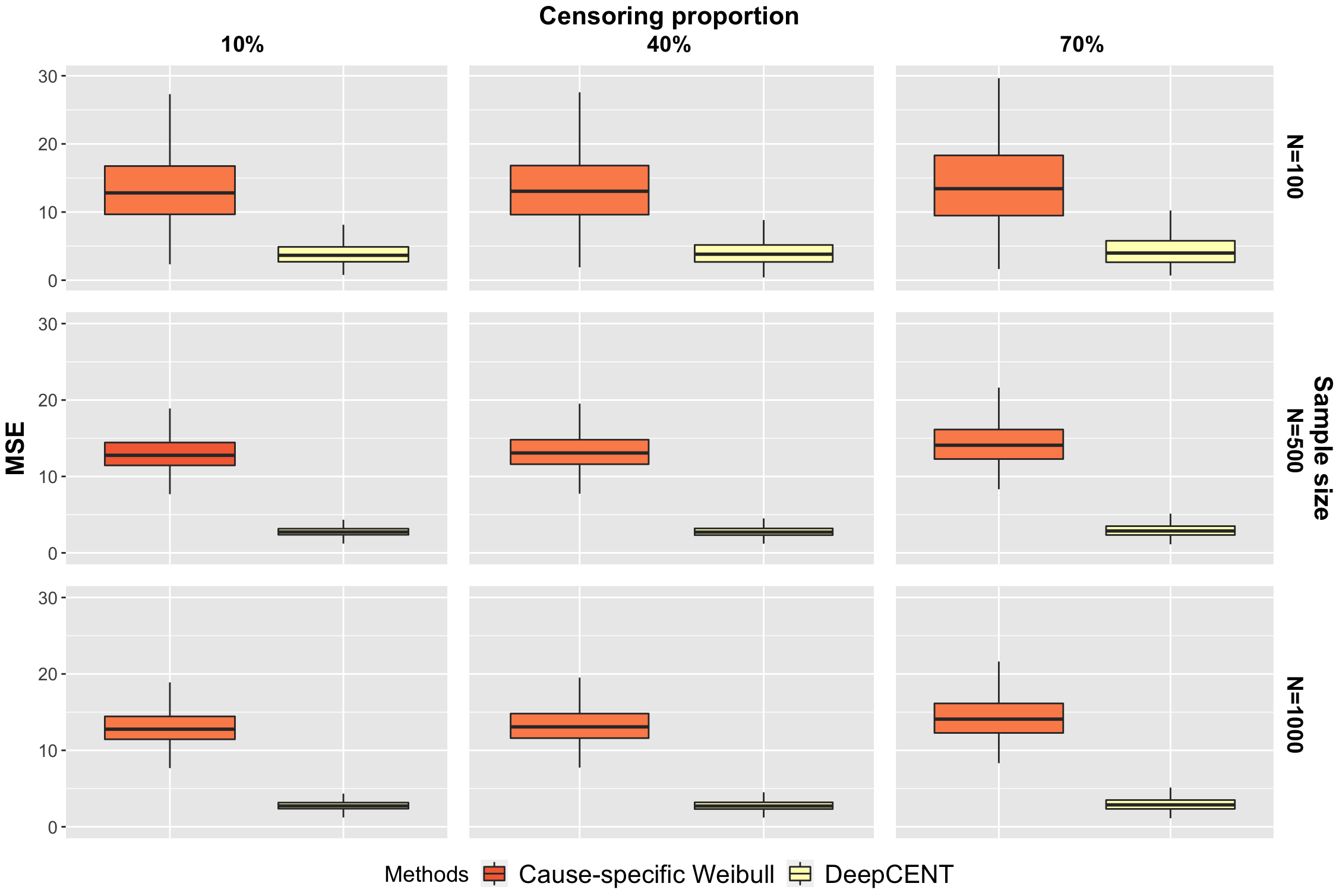}
  \caption{MSE for event type 1}
\end{subfigure}

\begin{subfigure}{0.49\textwidth}
  \centering
  \includegraphics[width=.95\linewidth]{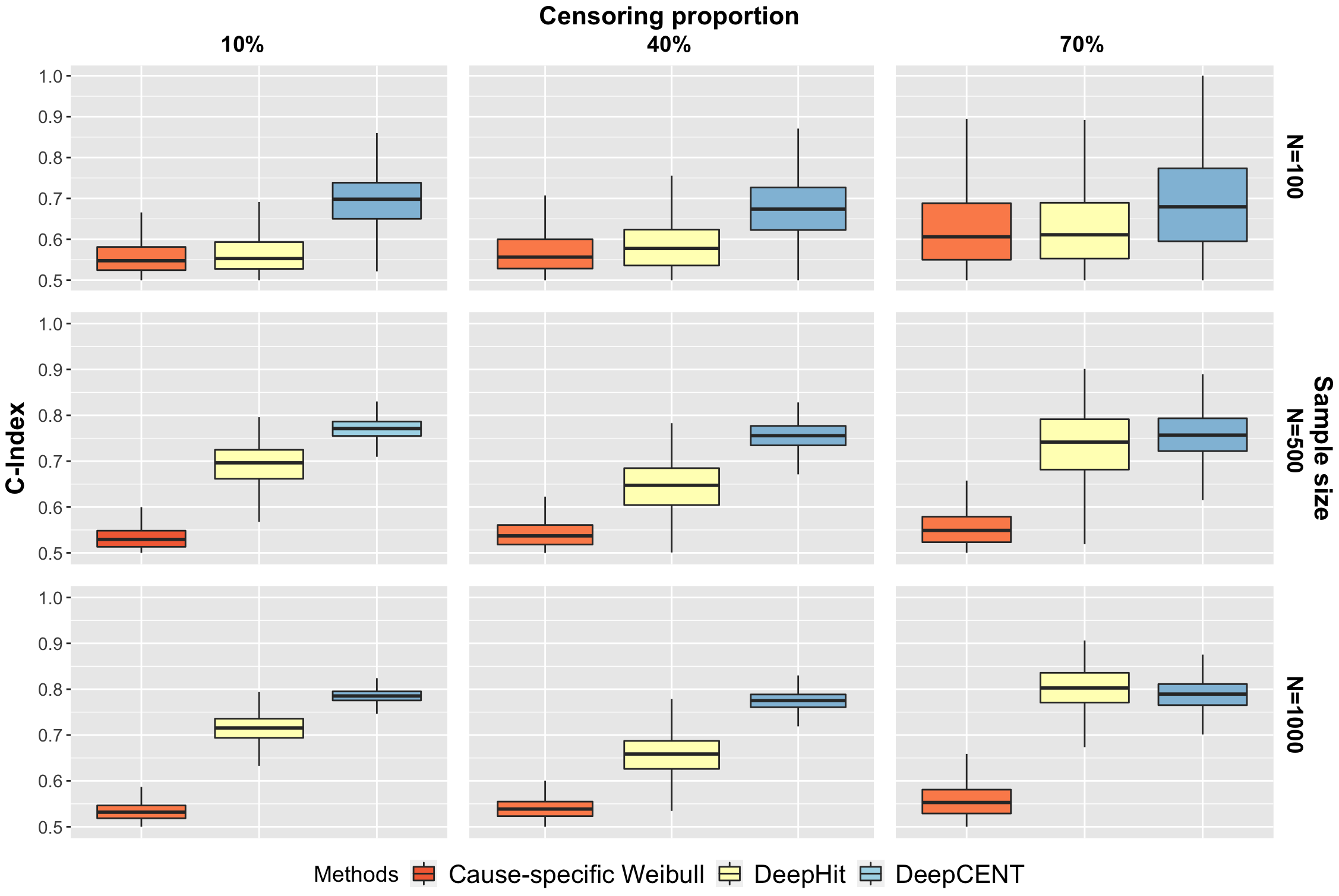}
  \caption{$C$-index for event type 2}
\end{subfigure}
\begin{subfigure}{0.49\textwidth}
  \centering
  \includegraphics[width=.95\linewidth]{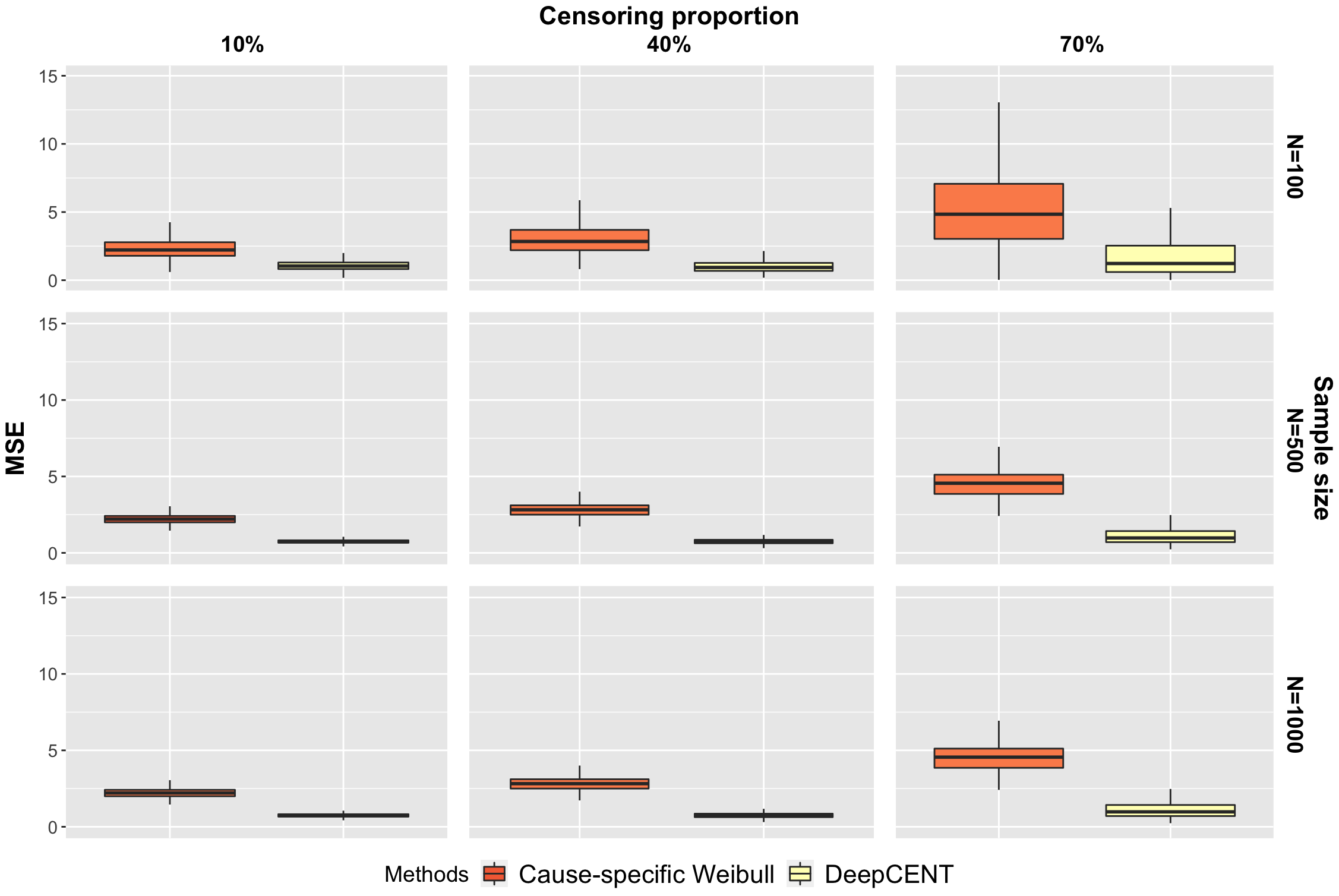}
  \caption{MSE for event type 2}
\end{subfigure}
\caption{Boxplots of $C$-index and MSE under competing risks}
\label{fig:simulation_cmprsk}
\end{figure}

Figure~\ref{fig:simulation_cmprsk} shows the results for the competing risks setting. For the $C$-index, DeepCENT has the highest $C$-index for all scenarios. DeepHit performs better then cause-specific Weibull regression, but it is not stable when the censoring rate is high (40\% or 70\%) especially for event type 1. The MSE between predicted and true event times is only calculated for Weibull and DeepCENT since DeepHit cannot predict event times. For the MSE, we observe that DeepCENT outperforms the Weibull regression in all scenarios as expected.

In both comparisons with or without competing risks, one can also observe that the variability of the C-index and the MSE tends to smaller with DeepCENT.

\section{Real Data Application}
In this section, we compared DeepCENT with the Weibull regression and RankDeepSurv/DeepHit using 3 real data sets: two noncompeting risks data sets and one competing risks data set.

\textbf{NKI70} The Netherlands Cancer Institute 70 gene signature data set \citep{van2002gene} contains 144 lymph node positive breast cancer patients' information on metastasis-free survival, 5 clinical risk factors, and 70 gene expression measurements found to be prognostic for metastasis-free survival in an earlier study. The censoring proportion is around 67\%. The covariates we included in the model are age at diagnosis, ER status, and 6 genes that show nonlinear relationship with the survival time (NUSAP1, FGF18, ZNF533, COL4A2, CDCA7, MCM6).

\textbf{METABRIC} The Molecular Taxonomy of Breast Cancer International Consortium \citep{curtis2012genomic} data set consists of 1981 breast cancer patients' gene expression data and 21 clinical features. We included top 50 genes that are marginally significant with survival time with other 11 clinical features. We restricted to the 1884 patients with complete data on the included covariates, where the censoring proportion is 42.14\%.

\textbf{Bladder} This is a publicly available data from bladder cancer patients with 1381 custom platform microarray features and other clinical information (GSE5479) \citep{dyrskjot2007gene}. The primary event is death due to bladder cancer and the competing event is death from other causes.
We included top 50 genes that are marginally significant with survival time with other clinical features (age, country, stage and grade). We restricted to the 183 patients with complete data on the included covariates, and there are 21.9\% primary events, 18.6\% competing events and 59.5\% censoring.

All of the data sets were randomly splitted into $\frac{2}{3}$ training and $\frac{1}{3}$ test sets and repeated for 50 times to calculate the average performance based on the $C$ -index and MSE. Note that since we do not know the true failure time in the real data setting, the MSEs are only calculated over the observed events.

Table \ref{tab:regular} shows that under noncompeting risks data, DeepCENT performs similar as RankDeepSurv and outperforms the Weibull regression with NIK70 data in both metrics. It performs slightly better than RankDeepSurv with the METARBIC data set. Interestingly, both of the deep learning methods did not outperform the Weibull regression with respect to the $C$-index with METBARIC data, although the larger MSEs indicate poor prediction ability by the Weibull regression. For competing risks data, DeepCENT outperforms the other two methods as expected. In these real data analyses, one can also observe that the variability of the C-index and the MSE tends to smaller with DeepCENT.

Taking the NKI70 data set as an example, for an ER positive patient who is diagnosed with breast cancer at 39 years old with gene expression level for those 6 genes at (-0.466, 0.432, -0.109, 0.267, -0.185, 0.060), the predicted time to metastasis is 1.25 with 95\% prediction interval (PI) = (1.40, 3.43) years, which measures the model uncertainty.

\begin{table}[h]
\centering
\caption{$C$-index and MSE (mean(SD)) of the real data application}
\label{tab:regular}
\begin{adjustbox}{width=0.8\textwidth}
\begin{tabular}{ccccc}

\hline
             & \multicolumn{2}{c}{NKI70} & \multicolumn{2}{c}{METABRIC}\\

             & $C$-index       & MSE             & $C$-index       & MSE           \\
\hline
Weibull      & 0.600 (0.105) & 16.030 (15.515) & 0.687 (0.018) & 1.399 (0.170) \\
DeepCENT   & 0.676 (0.064) & 0.721 (0.245)   & 0.633 (0.020) & 0.850 (0.120) \\
RankDeepSurv & 0.678 (0.058) & 0.731 (0.264)   & 0.622 (0.021) & 0.906 (0.133)\\
\hline
\hline
         & \multicolumn{4}{c}{Bladder} \\ 
         & $C$-index 1     & $C$-index 2     & MSE1             & MSE2   \\          
\hline
Weibull  & 0.643 (0.116) & 0.551 (0.139) & 731.99 (2254.10) & 418.08 \\ 
DeepCENT & 0.701 (0.080) & 0.635 (0.107) & 2.371 (1.127)    & 2.879 (2.196)\\
DeepHit  & 0.698 (0.138) & 0.649 (0.130) & NA               & NA    \\           
\hline

\end{tabular}
\end{adjustbox}
\end{table}

\section{Discussion}
In this paper, we proposed a method to predict individual survival times that utilize the deep learning framework with an innovative loss function. The newly proposed loss function consists two parts: a modified mean square error loss and a modified $C$-index ranking loss to handle right censored data, which not only consider the prediction accuracy but also the discriminative performance of the trained algorithm. A novel deep learning algorithm to jointly predict times to competing events was also developed. The results from simulation studies and real data applications confirmed that DeepCENT outperforms existing methods and can provide an effective solution to predicting individual survival times accurately with or without competing risks. Future research can merit incorporation of time-varying covariates or Autoencoder layers for dimension reduction in order to handle high-dimensional data.

\section*{Declarations of interest}
None

\section*{Acknowledgments}
This research was supported in part by the University of Pittsburgh Center for Research Computing through the resources provided.

\newpage

\bibliographystyle{biom}
\bibliography{mybib}

\newpage

\end{document}